\documentclass[runningheads]{llncs}
\usepackage{cite}
\usepackage{amsmath,amssymb,amsfonts}
\usepackage{algorithmic}
\usepackage{algorithm}
\usepackage{graphicx}
\usepackage{url}
\usepackage{textcomp}
\usepackage{xcolor}
\usepackage{subfig}
\usepackage{float}
\usepackage{bm}
\usepackage{geometry}
\begin{document}
\title{Decomposed Soft Actor Critic Method for Cooperative Multi-Agent Reinforcement Learning}%\thanks{Supported by organization x.}}
%
%\titlerunning{mSAC}
\author{Yuan Pu\inst{1}\and
Shaochen Wang\inst{1} \and
Rui Yang\inst{1}\and
Xin Yao\inst{1}\and
Bin Li \inst{1} }

%\authorrunning{Y. Pu et al.}
%% First names are abbreviated in the running head.
%% If there are more than two authors, 'et al.' is used.
%%
\institute{University of Science and Technology of China, Hefei, China \\
\email{\{puyuan, samwang, yr0013, xinyao\}@mail.ustc.edu.cn,
	binli@ustc.edu.cn}}

\maketitle              % typeset the header of the contribution
\begin{abstract}
Deep reinforcement learning methods have shown great performance on many challenging cooperative multi-agent tasks. Two main promising research directions are multi-agent value function decomposition and multi-agent policy gradients. In this paper, we propose a new decomposed multi-agent soft actor-critic (mSAC) method, which effectively combines
the advantages of the aforementioned two methods.
%multi-agent value function decomposition and policy-based method.
The main modules include decomposed Q network architecture, discrete probabilistic policy and counterfactual advantage function (optinal).
% which incorporates the idea of the multi-agent value function decomposition and soft policy iteration framework effectively and is a combination of novel and existing techniques, including decomposed Q value network architecture, decentralized probabilistic policy, and counterfactual advantage function (optional). 
Theoretically, mSAC supports efficient off-policy learning and addresses credit assignment problem partially in both discrete and continuous action spaces. Tested on StarCraft II micromanagement cooperative multiagent benchmark, we empirically investigate the performance of mSAC against its variants and analyze the effects of the different components. Experimental results demonstrate that mSAC significantly outperforms policy-based approach COMA, and achieves competitive results with SOTA value-based approach Qmix on most tasks in terms of asymptotic perfomance metric. In addition, mSAC achieves pretty good results on large action space tasks, such as $2c\_vs\_64zg$ and $MMM2$.

\end{abstract}
\section{Introduction}
Many real-world tasks can be modeled as multi-agent systems. Developing AI system for playing multi-agent games has raised much
attention. 
Recent years, deep multi-agent reinforcement learning (MARL) algorithms \cite{b1} have presented impressive results in many challenging multi-agent systems, such as the coordination of autonomous vehicles \cite{b2}, the challenging StarCraft II game \cite{b3}, etc. Maybe the simplest way to solve multi-agent system problems is, treating everything else as the environment for each individual agent, and learning concurrently based on the global reward. However this will face the issues \cite{b4}: (1) non-stationarity: when an agent is learning, the policies of other agents are also changing simultaneously, which means that the dynamic of environments is non-stationary; (2) scalability:
the joint state and action space grows exponentially as the number of agents increases.
To cope with these issues, 
most recent advanced algorithms adopted the paradigm of centralized training with decentralized execution (CTDE)\cite{b5}, in which they learn a centralized critic conditioned on joint action and observation history and take decentralized execution by learning different local actor (value functions or policies) for each individual agents. 
% By using centralized critic, the number of networks parameters is linearly proportional to the number of agents rather than

Following CTDE paradigm, there are two main popular and promising research lines in MARL, one is the value function decomposition approach, another is multi-agent policy gradients. Value Decomposition Network (VDN) \cite{b6} represented joint Q value $Q^{tot}$ as a sum of individual Q-values $q^{i}$ that condition only on individual actions and observations. Each decentralized policy arise simply from its local Q values $q^{i}$ (selects actions greedily by $q^{i}$). Afterwards, QMIX \cite{b7} employed a network to estimate joint action-values as a non-linear combination of per-agent values that condition on local observations. The representative work of multi-agent policy gradient method is COMA \cite{b8} method, which explicitly used a counterfactual baseline to address the challenges of multi-agent credit assignment and a critic representation to compute the counterfactual baseline efficiently.

Recent work \cite{b16} points out that multi-agent Q-learning with linear value decomposition implicitly implements a classical multi-agent credit assignment method called counterfactual difference rewards, which draws a connection with COMA. 
% However, value function decomposition is hard to apply in continuous domains 
However, value function decomposition is hard to apply in off-policy training and potentially suffers from the risk of unbounded divergence.
% So it's often necessary to resort to actor-critic methods.
%Atari 2600 arcade games \cite{b2}, board games \cite{b3}, robot manipulation tasks \cite{b4} \cite{b5} . 
In single-agent problems, to achieve sample efficiency and robust performance, \cite{b6} proposed the soft actor-critic algorithm, which is an off-policy actor-critic RL algorithm based on the maximum entropy reinforcement learning framework and achieves state-of-the-art performance on many challenging continuous control benchmarks. 

To attain both stability and good final performance in CTDE paradigm, how to effectively incorporate soft actor critic paradigm with multi-agent value function decomposition would be important. Following the research line of \cite{b14}, our key insight is, to efficiently compute the expected joint Q values, only when this linear condition --- the joint Q value $Q^{tot}$ is the linear mixture of the individual Q value ${q^{i}}$ satisfy, the following equation holds (Detailed proof can be found in Appendix), 
\begin{align}
	\mathbb{E}_{\boldsymbol{\pi}}\left[Q^{t o t}(\boldsymbol{s},\boldsymbol{\tau}, \boldsymbol{a})\right]
	&=\sum_{i} k^{i}(\boldsymbol{s}) \mathbb{E}_{{\pi}^{i}} \left[ q^{i}\left(\boldsymbol{\tau}^{i}, a^{i}\right)\right]+ b(\boldsymbol{s})\\
	&=q^{mix}(\boldsymbol{s},\mathbb{E}_{{\pi}^{i}} \left[ q^{i}\left(\boldsymbol{\tau}^{i}, a^{i}\right)\right])
\end{align}
where the  $Q^{tot}$ represents neural networks that consist of agent networks  ${q^{i}}$ and the mixing network ${q^{mix}}$. Note that, in our method, to make the aforementioned equation holds, the mixing network ${q^{mix}}$ is not a complex non-linear way but linear weights that is produced by the hyper-network only conditioned on global state information. 
% that ensures consistency between the centralised and decentralised policies.
% To this end, we restrict the q mix network as the linear function of global state information to  satisfy the aforementioned assumption.

Motivated by these insights, in this paper, we present a novel multi-agent soft actor-critic (mSAC) method, which is based on the following assumption: the joint Q value $Q^{tot}$ is the linear mixture of the individual Q values ${q^{i}}$. mCSAC contains three main components: decomposed soft Q network architecture, decentralized probabilistic policy, and counterfactual advantage function. This method incorporates the idea of the soft actor-critic and multi-agent value function decomposition effectively.
%The actor reuses past experience through off-policy optimization and acts maximize expected reward while also maximizing entropy to %increase exploration.

We empirically investigate the performance of our algorithm mSAC and analyze the influence of these components by ablation studies in StarCraft II micromanagement cooperative multi-agent tasks. Experiment results demonstrate that mSAC significantly outperforms current advanced policy-based algorithms (e.g. COMA) and achieves comparable performance with value-based approaches (e.g. Qmix) on most tasks. In addition, the variant method mSAC achieves pretty good results in large action space tasks, like \text {\em 2c\_vs\_64zg} and $MMM2$ task.
%and thereby improves performance even on domains that are fully observable and do not obviously require memory

To sum up, here are our contributions: 
\begin{itemize}
	\item We propose the novel mSAC method to effectively incorporate soft actor critic with value function decomposition method and investigate its practical performance on StarCraft II cooperative multi-agent benchmark. 
	\item We conduct extensive performance test of different mSAC variants to show the effect of soft value iteration, counterfactual advantage function, probabilistic policy, respectively.
\end{itemize} 

%Our code is available at https://github.com/puyuan1996/MARL.
%\footnote{\url{https://github.com/katerakelly/oyster}}.

\section{Related Works}\label{related work}
\subsection{Soft Actor-Critic}
Before introducing the Soft Actor-Critic method (SAC) \cite{b11}, we first briefly present the deep reinforcement learning (RL) problem definition. RL problem is often formulated as a {\em Markov Decision Process} (MDP), $\mathcal{M}=\left(\mathcal{S}, \mathcal{A}, p, r, \gamma\right)$. When the RL agent interacting with the environment, at each step, the agent observes a state ${\mathbf{s}_{t} \in \mathcal{S}}$, where $\mathcal{S}$ is the state space, and chooses an action ${\mathbf{a}_{t}\in\mathcal{A}}$, according to the policy $\pi(\mathbf{a}_{t}|\mathbf{s}_{t})$, where $\mathcal{A}$ is the state space, then the agent receives a reward $r\left(\mathbf{s}_{t}, \mathbf{a}_{t}\right)$ and the environment transforms to a next state $\mathbf{s}_{t+1}\sim p(\mathbf{s}_{t+1}|\mathbf{s}_{t},\mathbf{a}_{t})$. 
% Note that the environment's dynamics is contained in both the transition probabilities $p$ and reward functions $r$, which are usually unknown to the agent.%according to the reward function $r$,

% The objective of maximum entropy reinforcement learning framework is to maximize the discounted expected total reward plus the expected entropy of the policy:
The objective of reinforcement learning is to maximize the discounted expected total reward. However, in a maximum entropy RL framework, the goal is not only  to optimize the cumulative expected rewards, but also maximizes the expected entropy of the policy:
\begin{equation}
	J(\pi)=\sum_{t=0}^{T} \mathbf{E}_{\left(\mathbf{s}_{t}, \mathbf{a}_{t}\right) \sim \rho_{\pi}}\left[r\left(\mathbf{s}_{t}, \mathbf{a}_{t}\right)+\alpha \mathcal{H}\left(\pi\left(\cdot | \mathbf{s}_{t}\right)\right)\right]
\end{equation}
, where $\gamma $ is the discounted factor, $\rho_{\pi}\left(\mathbf{s}_{t},\mathbf{a}_{t}\right)$ denotes 
% the state-action marginal of the trajectory distribution 
the state-action marginal distribution of the trajectory
induced by the policy $\pi(\mathbf{a}_{t}|\mathbf{s}_{t})$.
% {\em Soft Actor-Critic} 
% is an  off-policy algorithm and 
SAC is a popular single-agent off-policy actor-critic method
using the maximum entropy reinforcement learning framework. It utilizes an actor-critic architecture with separate policy and value networks, an off-policy paradigm that enables reuse of previously collected data, and entropy maximization to enable effective exploration. In contrast to other off-policy algorithms, SAC is quite stable and has been considered as a state-of-the-art baseline for a diverse range of RL problems with continuous actions.
% achieving state-of-the-art results on a range of continuous control benchmarks. 

\subsection{Value Function Decomposition}
Value function decomposition (VDN ) \cite{b9} methods learn local Q value functions for each individual agent, and then these local Q values are combined with a learnable mixing neural network to produce
joint Q values. 
\begin{equation}
	Q^{t o t}(\tau, \mathbf{a})=q^{mix}(\boldsymbol{s}, \left[ q^{i}\left(\boldsymbol{\tau}^{i}, a^{i}\right)\right])
\end{equation}
In VDN, the mixing function ${ q^{mix}}$ is a simple algorithmic summation. While in QMIX, it's a non-linear monotonic factorization structure, which can achieve a much richer function class at the the same time satisfy the principle of the Individual-Global Maximization (IGM):  a global $argmax$ performed on $Q^{tot}$ yields the same result as a set of individual $argmax$ operations
performed on each local $q^i$.

\subsection{Multi-Agent Policy Gradients}
The centralized training with decentralized execution (CTDE)
paradigm has recently attracted attention for its ability
to address non-stationarity problems 
% while maintaining decentralized execution. 
Learning a centralized
critic with decentralized actors (CCDA) is an efficient approach that exploits the CTDE paradigm.
COMA and MADDPG are two representative examples.

COMA uses a centralised critic to estimate the Q function
and decentralised actors to optimise the agents’ policies. To address the challenges of multi-agent credit assignment,
it uses a counterfactual baseline that marginalises
out a single agent’s action, while keeping the other agents’
actions fixed. In addition, COMA also uses a critic representation that allows
the counterfactual baseline to be computed efficiently in
a single forward pass. And it updates stochastic policies using the gradients:
\begin{equation}
	g=\mathbb{E}_{\boldsymbol{\pi}}\left[\sum_{i} \nabla_{\theta_{i}} \log \pi^{i}\left(a^{i} \mid \tau^{i}\right) A^{i}(\tau, \boldsymbol{a})\right]
\end{equation}
where, $$
A^{i}(\tau, a)=Q^{t o t}(\tau, a)- \sum_{\boldsymbol{a^{\prime,i}}} {\pi}^{i}({a^{\prime,i}} \mid \boldsymbol{\tau}^{i}) Q^{t o t}\left(\tau,\left(a^{-i}, a^{\prime,i}\right)\right)$$  is a counterfactual advantage and ${a_{-i}}$ is the
joint action other than agent i.

MADDPG\cite{b10} is an adaptation of actor-critic methods which learns deterministic policies in continuous action spaces, considers action policies of other agents and is able to successfully learn policies that require complex multi-agent coordination.

%To verify our hypothesis
%that latent context could increase the the agent's memory length and implicitly benefit the representation learning, our modification is based on SAC.

\section{Methods}

%Our method is especially suitable for long time-horizon decision problems in which the agent must %memorize and leverage recent experiences to effectively solve problems. 
In this section, we first introduce the definition and notation of the Decentralized Partially Observable Markov Decision Process (Dec-POMDP) and then introduce the multi-agent policy gradient decomposed architecture. Afterwards we present the three variant method: multi-agent soft actor-critic (mSAC) method, multi-agent counterfactual soft actor-critic (mCSAC) method, multi-agent counterfactual actor-critic (mCAC) method, respectively.
\subsection{Problem Formulation}
The fully cooperative multi-agent tasks can be modelled as a decentralized partially observable Markov decision process (Dec-POMDP) \cite{b12} $$G=\langle S, A, P, R, \Omega, O, n, \gamma\rangle$$, 
%  I is the finite set of agents, 
where $s \in S$ is the global state and 
% $\gamma$ is the discount factor. 
$o \in \Omega$ is a local observation.
At each time-step, each agent i
receives an observation $ o^i$ drawn according to the observation function 
$O(s, i)$ and selects an action $a^{i} \in A^{i}$, forming a joint action $\boldsymbol{a} \in {A} \equiv {A^n} $, and the environment transitions to the next state $s^{\prime}$ according to the
transition function $P\left(s^{\prime} \mid s, a\right)$ and receiving a reward $r=R(s, a)$ shared by all agents. Each agent learns a
policy $\pi^{i}\left(a^{i} \mid \tau^{i} ; \theta_{i}\right)$, which is parameterized by $\theta_{i}$ and conditioned on the local observation-action history $\tau^{i} \in \mathrm{T} \equiv
(\Omega \times A)^{*}$. The joint policy $\boldsymbol{\pi}$, with parameters $\theta=\left\langle\theta_{1}, \cdots, \theta_{n}\right\rangle$, constitute of the joint Q function:
$	Q_{\pi}^{ t o t}(\tau, a)=\mathbb{E}_{s_{0: \infty}, a_{0: \infty}}\left[\sum_{t=0}^{\infty} \gamma^{t} R\left(s_{t}, a_{t}\right) \mid s_{0}=s, a_{0}=a, \pi\right]$
%\begin{small}
%	\begin{equation}
%		Q_{\pi}^{ t o t}(\tau, a)=\mathbb{E}_{s_{0: \infty}, a_{0: \infty}}\left[\sum_{t=0}^{\infty} \gamma^{t} R\left(s_{t}, a_{t}\right) \mid s_{0}=s, a_{0}=a, \pi\right]
%	\end{equation}
%\end{small}
\subsection{Multi-Agent Decomposed Policy Gradient Architecture}
In this part, 
% before introducing the concrent counterfactual multi-agent soft actor-critic method, 
we first present the common multi-agent decomposed policy gradient architecture for all the algorithm variants we will introduce in the next three subsections.
% , we present the there algorithm variants, respectively.
% followed \cite{b11} , 

We use function approximators (neural networks) for both the centralized critic: Q-function and the decentralized actor: policy, and alternate between optimizing both networks with stochastic gradient descent. We will consider a parameterized Q function $Q_{\phi}(s_t,\tau_t,a_t)$ and a tractable policy $\pi_{\theta}(a_t | \tau_t)$., where ${\phi}$ and ${\theta}$ are referred to the parameters of the Q networks and policy networks, respectively.

{\bf Policy Network} Also called decentralized actor. For simplicity,
our decentralized actor (policy) network structure is the same as the agent i's local Q network except that one $clamp (-5,2)$ operation and a $softmax$ layer is added after the local Q network. At the beginning of training, if the policy networks have improper initialization parameters which would result in policy distribution becomes too sharp potentially, and thereby constrain the degree of exploration. Empirically, we found this clamp operation relieves this issue and accelerates training. The softmax layer is to convert probabilistic logits to the categorical distribution. 
The policy network parameter is shared among all agents, and different agents are distinguished by utilizing a one-hot identity vector, in order to be consistent with Qmix and fair comparing. 
% of course we can use different policy network for different individual agent.

{\bf Value Network} The centralized critic's network structure is modified from Qmix's Q network structure, and is comprised of agent i's local Q network $q^{i}$ and mixing network $q^{mix}$ in which the weights and biases are produced by the separate hyper-networks. Figure 1 illustrates detailed structure of local Q network and the mixing network. 
For each agent i, there is one local Q network that represents its local Q value function $q^{i}(\tau^i, a^i)$. We represent local Q networks as GRUs \cite{b17} that receive the current individual
observation $o^i_t$ and the last action $a^i_{t-1}$ as input at each time step.
The mixing network is a feed-forward neural network that
takes the agent's local Q network outputs as input and mixes them
linearly and followed by an absolute activation function, producing the values of $Q^{tot}$, as shown in Figure 1. To make the equation (1) holds,
% the linear condition above of (5), 
the weights and the biases of the mixing network are restricted
to be linear functions of $s$, and the parameters are produced by the separate
hyper-networks same as in Qmix,
which allows us to effectively calculate the expected Q values.
% The weights and the biases of the mixing network.

\begin{figure}[!t]
	\centering
	\includegraphics[width=8cm,height=4cm]{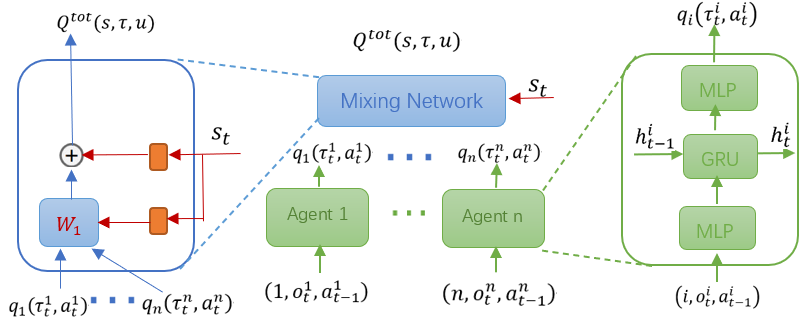}
	\caption{Left: mixing network structure. Red figures are the hyper-networks that produce the weights and biases for the mixing network layers. 
		% 	are shown in blue. 
		Middle: the overall Qmix architecture. 
		Right: agent's local Q network, which is in green, the $i$ means the corresponding one-hot vector to distinguish different agents.}
\end{figure}
\subsection{multi-agent Soft Actor-Critic (mSAC)}
\begin{algorithm}[tb]
	\caption{mSAC}
	\label{alg:algorithm}
	Initialize network parameters: $\theta, \phi_{1,2}, \bar{\phi}_{1,2}$, 
	and replay buffers: $\mathcal{D}$ for training  policy and value networks. 
	max training episodes$M$, replay buffer size $rlbs$
	\begin{algorithmic}[1] %[1] enables line numbers
		\FOR{episode = $1$ to $M$ }%each collecting data step t}{ 
		\STATE for each agent i, observe the global state s and its individual observations  $o^i$\;
		\FOR{t = 1 to max-episode-length }
		\STATE for each agent i, , select action $\mathbf{a}_t^i$ according to the current policy $\pi_{\theta}(\mathbf{a}_t^i | \tau_t^i)$ 
		\STATE Execute actions$\mathbf{a}=(a^1,a^2,...,a^N)$ and interact with the environment and obtain the global reward$r$, and the environment transitions to the next global state $s'$
		\STATE add the experience $ \left( \mathbf{s}_{t},{o}^{i}_{t}, \mathbf{a}_{t},\mathbf{r}_{t}, \mathbf{s}_{t+1},{o}^{i}_{t+1} \right)$ to the replay buffer $\mathcal{D}$
		\FOR{each rl training step}
		\STATE Sample a random minibatch of  $\mathbf{B}$ uniformaly from  $\mathcal{D}$
		\STATE Update critic network w.r.t the equation (6) \\
		$\phi_{i} \leftarrow \phi_{i}-\alpha_{Q} \hat{\nabla}_{\phi_{i}} J_{Q}\left(\phi_{i}\right)$, \textbf{for} $i \in\{1,2\}$
		\STATE Update actor network  w.r.t the equation (9) \\%using the sampled policy gradient
		$\theta \leftarrow \theta-\alpha_{\pi} \hat{\nabla}_{\theta} J_{\pi}(\theta)$
		\STATE Update hyper-parameter$\alpha$\ w.r.t the equation (11) \\
		Update target value network parameters for each agent i:\\
		$\bar{\phi}_{i} \leftarrow \tau \phi_{i}+(1-\tau) \bar{\phi}_{i}$, for $i \in\{1,2\} $
		\ENDFOR
		\ENDFOR
		\ENDFOR
		\STATE \textbf{return} $Q_{\phi},\pi_{\theta}$  
	\end{algorithmic}
\end{algorithm}

Before introducing the counterfactual multi-agent soft actor-critic method, 
% followed \cite{b11} , 
we first present multi-agent soft actor-critic method (we refer it as mSAC), which adopts the practical approximation to soft policy iteration as in \cite{b11}.

Similar with \cite{b11}, the critic loss function of the mSAC method in multi-agent setting is,
\begin{equation}
	\mathcal{L}(\phi)=\mathbb{E}_{ \mathcal{D}}\left[\left(r_{t}+\gamma * \min_{j \in {{1,2}}} \hat{Q}_{\phi_j^{\prime}}^{targ}-Q^{t o t}_{\phi}\left(\boldsymbol{s}_{t},\boldsymbol{\tau}_{t}, \boldsymbol{a}_{t}\right)\right)^{2}\right]
\end{equation}%
Based on original soft actor-critic algorithm, our methods also utilize two soft Q-value networks  $Q^{t o t}_{\phi_j}\left(\boldsymbol{s},\boldsymbol{\tau}, \boldsymbol{a} \right)$, for $j \in\{1,2\}$, and take the min values as the target.
In equation (7),
%\begin{align}
%    \hat{Q}^{t o t}=r_{t}+\gamma * \min_{i} \hat{Q}_{i}^{targ}
%\end{align}
\begin{small}
	\begin{align}
		\hat{Q}_{\phi_j^{\prime}}^{targ}&=\mathbb{E}_{\boldsymbol{\pi}_{\theta}}\left[ Q^{t o t}_{\phi_j^{\prime}}\left(\boldsymbol{s}_{t+1},\boldsymbol{\tau}_{t+1}, \boldsymbol{a}_{t+1} \right)-\alpha \log \boldsymbol{\pi}\left(\boldsymbol{a}_{t+1} \mid \boldsymbol{\tau}_{t+1}\right)\right]\\
		&=q^{mix}\left(\boldsymbol{s}_{t+1},\mathbb{E}_{{\pi}^{i}} \left[ q^{i}\left(\boldsymbol{\tau}^{i}_{t+1}, a^{\prime, i}_{t+1}\right)-\alpha \log \pi^{i}\left(a^{\prime, i}_{t+1} \mid \tau^{i}_{t+1}\right)\right]\right)
	\end{align}
\end{small}
and $\mathcal{D}$ is a replay buffer containing previously sampled transitions (states, local observations, actions, rewards, next states, next local observations): 
%$\left(\boldsymbol{s}_{t},\tau_{t}, \boldsymbol{a}_{t},r_{t},\boldsymbol{s}_{t+1}, \boldsymbol{\tau}_{t+1}\right) $,
 and ${Q}^{tot}_{{\phi}_{j^{\prime}}}$  is the target Q network, with parameters ${\phi}_{j^{\prime}}$ that are obtained as an exponentially moving average of the current Q network weights  ${\phi}_{j}$, which has been shown to stabilize training.  

Note that, in equation (9), $a^{\prime, i}_{t+1}$ is sampled from agent i's current policy $\pi^i$ rather than sampled from the replay buffer. Compared with Qmix, we adopted the additional policy network that outputs the probabilistic policy (which is a probability mass function for discrete domain), which exactly represents the probabilistic value of each agent selects each discrete action. therefore we can calculate the expectation values exactly.

%a policy network $\pi_{\phi}\left(\mathbf{a}_{t} | \mathbf{\tau}_{t} \right)$. We also adopted the soft policy iteration %framework, alternating between optimizing these networks with stochastic gradient descent.

Recent work theoretically proved that the the soft (or call Boltzmann) policy iteration is guaranteed to improve and can converge to the optimal policy. Derived from the soft policy iteration procedure, the objective for policy update is below: 
%actor loss function is,
\begin{align}
	\mathcal{L}(\theta)&= \mathbb{E}_{\mathcal{D}}\left[ \alpha \log \boldsymbol{\pi}\left(\boldsymbol{a}_{t} \mid \boldsymbol{\tau}_{t}\right)-Q^{t o t}_{\phi^{\prime}}\left(\boldsymbol{s}_{t},\boldsymbol{\tau}_{t}, \boldsymbol{a}_{t} \right) \right]\\
	&=q^{mix}(\boldsymbol{s}_{t},\mathbb{E}_{{\pi}^{i}} \left[ q^{i}\left(\boldsymbol{\tau}^{i}_{t}, a^{i}_{t}\right)-\alpha \log \pi^{i}\left(a^{i}_{t} \mid \tau^{i}_{t}\right)\right])
\end{align}
$\alpha$ is a hyper-parameter that controls the trade-off between maximizing the entropy of policy and the expected discounted return. 
%However, choosing the optimal $\alpha$ is non-trivial. 
%For the same reason as in \cite{b11}, we also modified the $\alpha$ automatically with the following objective:

However, $\alpha$ need to be set as different values  at different stages of training or on different tasks. Because in different states, the degree of exploration needed is different. In some states, good policy have been learned, and the corresponding $\alpha$ value should be reduced to very samll to weaken the degree of exploration, but in other states, it's not sure which action is good and which action is bad, so we need to increse the degree of exploration. The SAC algorithm proposes to reconstruct the original soft policy iterative process as a constrained optimization problem, that is, when optimizing the policy to maximize cumulative discount returns, the algorithm should keep the average entropy of policy a fixed value (usually $-|A|$) and the action entropy in different states can be variable.  Specifically, $\alpha$ is automatically updated by optimizing the following loss \cite{b11}:
% $	L(\alpha)=\mathbb{E}_{\mathbf{a}_{t} \sim \pi_{t}}\left[-\alpha \log \pi_{t}\left(\mathbf{a}_{t} \mid \mathbf{\tau}_{t}\right)-\alpha \overline{\mathcal{H}}\right]$
\begin{equation}
	L(\alpha)=\mathbb{E}_{\mathbf{a}_{t} \sim \pi_{t}}\left[-\alpha \log \pi_{t}\left(\mathbf{a}_{t} \mid \mathbf{\tau}_{t}\right)-\alpha \overline{\mathcal{H}}\right]
\end{equation}

The details of the mSAC algorithm are summarized in Algorithm 1.
% , and the overview training process is shown in Figure 2.

\subsection{multi-agent Counterfactual Soft Actor-Critic (mCSAC)}
One of the most important problems in multi-agent reinforcement learning is credit assignment. For partially solving this issue, we adopted the insight in COMA that is using the counterfactual advantage function when we optimize the individual policy in the multi-agent decomposed policy gradient paradigms. The loss function of policy in multi-agent  counterfactual soft actor-critic (mCSAC) method is as following:
\begin{align}
	\mathbb{E}_{\left(\boldsymbol{s}_{t},\tau_{t}, r_{t}, \right) \sim \mathcal{D},  \boldsymbol{a}_{t}\sim {\boldsymbol{\pi}_{\theta}}}\left[
	\log \pi^{a}\left(a^{i}_{t} \mid \tau^{i}_{t}\right)  A^{i}(s_t, \tau_{t},\mathbf{a}_{t})\right]
\end{align}
where,
\begin{small}
	\begin{align}
		A^{i}(s_t, \tau_{t},\mathbf{a}_{t})=&-\alpha \log \pi^{i}\left(a^{i}_{t} \mid \tau^{i}_{t}\right)+ Q^{t o t}_{\phi}\left(\boldsymbol{s}_{t},\boldsymbol{\tau}_{t}, \boldsymbol{a}_{t} \right)\\
		&-q^{mix}\left(\boldsymbol{s}_{t},\mathbb{E}_{{\pi}^{i}} \left[ q^{i}\left(\boldsymbol{\tau}^{i}_{t}, a^{i,{\prime}}_{t}\right)\right],
		q^{-i}\left(\boldsymbol{\tau}^{-i}_{t}, a^{-i}_{t}\right)
		\right)
		%Q^{t o t}_{\phi}\left(\boldsymbol{s}_{t},\boldsymbol{\tau}_{t}, %\boldsymbol{a}_{t}\right)\right)\right]
	\end{align}
\end{small}
% in equation (12)'s right hand, $\boldsymbol{a_t}=(a^i,a^{-i})$, $a^{i,{\prime}}_{t}$ is sampled from agent i's current policy $\pi^i$, $a^{-i}_{t}$ is sampled from the replay buffer $\mathcal{D}$, 

Note that on the right side of the above equation, $a_{t}=\left(a^{i}_t, a^{-i}_t\right)$, $a_{t}$ refers to the joint actions at time $t$, and $a^{i}_t$ refers to the local action of the agent $i$, and $a^{-i}_t$ refers to the (partial) joint actions of the agents other than the agent i. $a_{t}^{-i}$ is sampled from the current policy $\pi_i$ of agent i, and $a_{t}^{-i}$ is sampled from the replay buffer $\mathcal{D}$ ,
$q^{mix}\left(.\right)$  calculates the counterfactual baseline, which measures the expected action value under the individual policy of the agent $i$ when fix the actions of other agents except agent $i$. If the joint action value of the samples sampled from the replay buffer 
$\left( \boldsymbol{s}_{t},\boldsymbol{\tau}_{t}, \boldsymbol{a}_{t} \right)$ 
is greater than the previous baseline, then we update the policy network parameters of the agent $i$ to increase the action probability of $a^{i}_t$, and vice versa.

% We summarize an overview of mCSAC approach in Appendix Algorithm 1. 
% sample $\mathbf{c} \sim q_{\omega}\left(\mathbf{c} | \mathbf{e}\right)$

% and found that it will harm the final performance to some extent, and obviously including %$\mathbf{s}_{t}$ in $\mathbf{e}$ will increase the computation cost. So in our most experiments, we %do not use the next observation of $\mathbf{e}$.

\subsection{multi-agent Counterfactual Actor-Critic (mCAC)}
In order to probe the effect of the soft policy iteration paradigm on multi-agent policy optimization, in this section we introduce another variant of the mSAC method: the multi-agent counterfactual actor critic (mCAC) method. 
which can be obatained after deletes the entropy augment item corresponding to $\alpha log \pi$ from all loss functions of the mCSAC method. Because it does not satisfy the condition of soft policy iteration paradigm, mCAC becomes an on-policy algorithm. The capacity of the replay buffer used to update the policy and the value network is set to a small value.

The behavior strategy used to collect trajectory experience is not based on the categorical distribution that outputed by the policy network, but a strategy similar to $\epsilon$-greedy exploration.
%and our soft policy evaluation is
% In order to avoid repetition, we no longer write the actor and evaluator loss objective function here.
The specific implementation of mCAC is similar to the mSAC algorithm. 
Action probabilities are produced from the final layer, z, via a bounded softmax distribution that lower-bounds the probability of any given action by $\epsilon/|A|$:
$P(a) = (1-\epsilon)* softmax(z)_a + \epsilon/|A|)$. We anneal $\epsilon$ linearly
from 0.5 to 0.02 across 20000 training episodes.

% The final action probability of the policy network is the distribution generated from the last layer z through the bounded softmax, adding an ϵ/|A| to the probability of each action to limit any given action. The lower bound of probability guarantees certain exploration, namely
% $P(a)=(1-\epsilon) * \operatorname{softmax}(z)\_{a}+\epsilon /|A|$. After 20,000 rounds, ϵ linearly anneals from the initial value of 0.5 to 0.02, and then ϵ remains unchanged at 0.02.

% In this part, we introduce the multi-agent counterfactual actor-critic (we noted it as mCAC) method which is a on-policy algorithm, which is attained after simply removing the $\alpha \log \pi^{i}$ entropy augmented term in all expressions of mCSAC method. To avoid redundancy, we won’t write the critic and actor loss objective here again.

\section{Experiments}
In this section, we present our experimental results and some analysis.  First, we describe the decentralised cooperated StarCraft II micromanagement benchmark to which we apply our proposed method mSAC and the variant methods we consider. Then we present the performance comparison between mSAC, the ablation variant algorithms mCSAC, mCAC and the representative value decomposed algorithm---Qmix and policy gradient algorithm---COMA   
in aforementioned discrete action environments. 
We used the Qmix implementation from this open-source code \footnote{\url{https://github.com/starry-sky6688/StarCraft}} with the same hyper-parameters as \cite{b7}. 
To be consistent with previous work, our implementation \footnote{\url{https://github.com/puyuan1996/MARL}} 
almost use the same network architecture and hyper-parameters across all the tasks.
 More experimental details can be found in the Appendix.

{\bf Experimental Setup}
We focus on the StarCraft II decentralized micromanagement tasks \cite{b13} \footnote{We use StarCraft 2 Version SC2.4.10 in our experiments.}, in which each of the agents controls an individual army unit and each agent receives a global shared reward. We use StarCraft Multi-Agent Challenge (SMAC) environment [15] as our APIs, which has become
a common-used benchmark for evaluating state-of-the-art MARL approaches such as COMA, QMIX and other baseline algorithms. In this paper, our algorithm learns multiple agents (or called policies) to control allied units to beat the
enemy, while the enemy units are controlled by a built-in handcrafted AI, which make use of the handcrafted heuristics. 
Two representative StarCraftII micromanagement
scenarios ($3m$ and $2c\_vs\_64zg$) are shown in Figure 2.
\begin{figure}[h]
	\centering
	\includegraphics[width=0.5\textwidth]{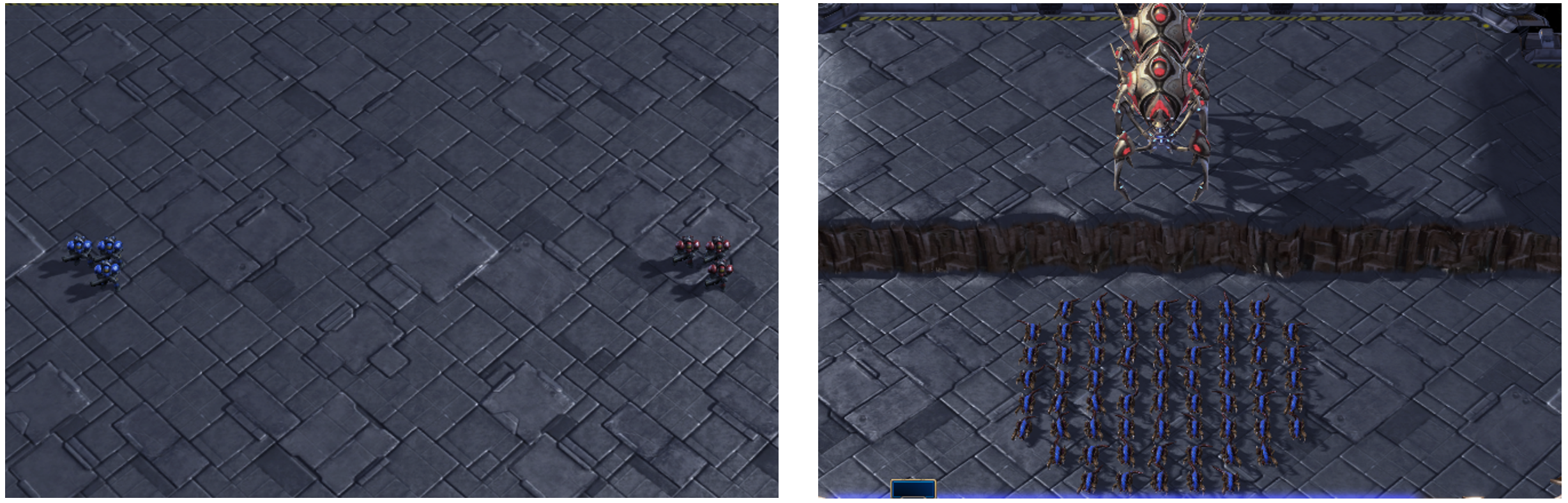} 
	%\subfloat[\centering]{{\includegraphics[width=5cm]{fig/map1.png} }}%
	\caption{Visualizations of the two representative StarCraftII micromanagement scenarios ($3m$ and $2c\_vs\_64zg$).}
	\label{wolf2}
\end{figure}

For comparing each method’s performance justly as much as possible, our Qmix implementation also use target Q networks that are obtained as an exponentially moving average of the Q function weights, which was different from the hard update manner in the original paper \cite{b7}. 
In addition, we adopt the same evaluation procedure as in  \cite{b7}. For each run of a method, we pause training every 100 episodes and run 20 independent episodes where each agent performing greedy decentralised action selection (for Qmix chosen the action with the largest local Q values, for other methods chosen the action with the largest probability value). The percentage of these episodes in which the method defeats all enemy units within the (different) time limit is referred as the test win rate.

% Figures 3 shows the performance (the mean test win rate) comparison across 5 runs for each method on the maps with homogeneous agents, Figures 4 shows the performance comparison for each method on the maps with heterogeneous agents and Figures 5 shows the performance comparison for mSAC and Qmix on large-action space map, such as $2c\_vs\_64zg$, $MMM2$ and $bane\_vs\_bane$.
The magnitude of x-axis is 100 episodes, and for different maps, there are different episode length limits according to the difficulty level of different tasks. The shaded region indicates the one  quarter of standard deviation.

{\bf Algorithm Details of Variant Methods}
The policy network of all agents includes a recurrent layer composed of GRUs with 64-dimensional hidden states, and a fully connected MLP layer before and after this. % Evenly sample 32 rounds of complete trajectory from the playback buffer, and then fully expand the trajectory for training through a cyclic neural network.
 % During training and testing, if the two armies are alive at the end of a round, we consider it a loss.
 After the team is defeated or the time step limit is reached, one episode ends.
 The mixed network part of the value function consists of a single hidden layer of 64 units, and the ELU nonlinear activation function is not used. Its weights and biases are generated by an additional hyper-network composed of a single hidden layer of 64 units without the ReLU nonlinear activation function.
 
 Similar to Qmix, the mSAC algorithm training is also carried out in mini-batch mode, the batch size is 32, the target smoothing coefficient used to update the two target Q networks is 0.005, and the discount coefficient is set to 0.95.
 Due to parameter sharing, all agents will be processed in parallel, and the information of each agent at each time step of each episode occupies one entry of the mini-batch. Once a new episode of trajectory is added to the replay buffer, the algorithm will update the network parameters of the actor and critic.
 Specifically, after collecting a episode of trajectory, 32 episodes were sampled from the replay buffer as a mini-batch to train the actor and the critic, fully expand the recurrent network part of the actor and the critic at all time steps and backpropagate the gradients, then apply the summarized gradient update to the neural network. For clarity, the hyperparameter settings of the mSAC algorithm are summarized in Table 1.
 
 \begin{table}[!htbp]
	\centering
	\caption{Hyper-parameters}
	\begin{tabular}{ll}		
		\hline
		Parameter Name   & Value  \\
		\hline
		leaning rate      & 5e-4     \\
		% 		entropy coefficient &0.05 \\
		target smoothing coefficient ($\tau$) & 0.005\\
		discount factor          & 0.99        \\
		optimizer         & RMSprop      \\
		activation function   & ReLU      \\
		replay buffer size (Off-policy)    & 5000 episodes   \\
		replay buffer size (On-policy)     & 32 episodes   \\
		%context batch size        & 100 or 20   \\
		RL batch size        & 32 episodes    \\		
		KL lambda & automated adoptiing \\
		entropy target & dim(A) (e.g. , -9 for 3m) \\
		
		% 		$N_{c+rl}$ & 1000 \\
		% 		{\em Horizon} & 100 or 20 \\
		% 		\hline
		% 		$N_{c}$ & 400 \\	
		% 		$N_{rl}$ & 600 \\	
		% 		training steps each iteration ($N_{train}$) & 4000 or 1000\\	
		\hline
	\end{tabular}
	\label{tab:booktabs}
\end{table}

  The learning performance of the mSAC method and its variant methods mCSAC and mCAC on the StarCraft II micro-operation task were tested separately to study the impact of off-policy update, soft Q value, probability distribution policy, counterfactual advantage function and other modules on the multi-agent policy gradient algorithm.
 In all maps, all algorithms used reward standardization techniques for stability purposes,
% We perform ablation variant experiments in order to investigate the influence of the necessity of soft Q values and the importance of counterfactual advantage function. 
% In all experiments, we use reward standardization for stability purpose,
\begin{equation}
	r_{standard} = 10 * (r-mean)/(r - std + 1e-6)
\end{equation}
% and the policy network structure is the same as the local q value network.
For clarity, we briefly outline the key differences of our different variant methods in Table 2.
\begin{table*}[!htbp]
	\centering
	\caption{Comparison of Variant Methods}
	\begin{tabular}{llllll}
		\hline
		method  & on/off policy  &  buffer size & counterfactual advantage function& soft Q values\\
		\hline
		mSAC &off-policy&5000 episodes&no&yes \\
		mCSAC &off-policy&5000 episodes&yes&yes \\
		mCAC &on-policy&32 episodes&yes &no \\	
		\hline
	\end{tabular}
	\label{tab:booktabs}
\end{table*}

% mCSAC (multi-agent counterfactual soft actor-critic): 
% 1. off policy, the actor and critic buffer size is set as 5000 episodes, which is same as in \cite{b7}; 2. using counterfactual advantage function; 3. using soft Q values. 

% mSAC (multi-agent soft actor-critic): 
% 1. off policy, the actor and critic buffer size is also set as 5000 episodes; 2. not using counterfactual advantage function; 3. using soft Q values.

% mCAC ({multi-agent counterfactual actor-critic):
% 1. on policy, the actor and critic buffer size is set as 32 episodes; 2. using counterfactual advantage function; 3. not using soft Q values; 4. As in \cite{b15}, delayed policy updates: only updating the actor and target critic network every d iterations, with d = 2. 

% \subsection{Main results}
{\bf Experimental Results}
\begin{figure}[!t]%[htbp]
	\centering
	\subfloat[\centering $8m$]{{\includegraphics[width=5cm]{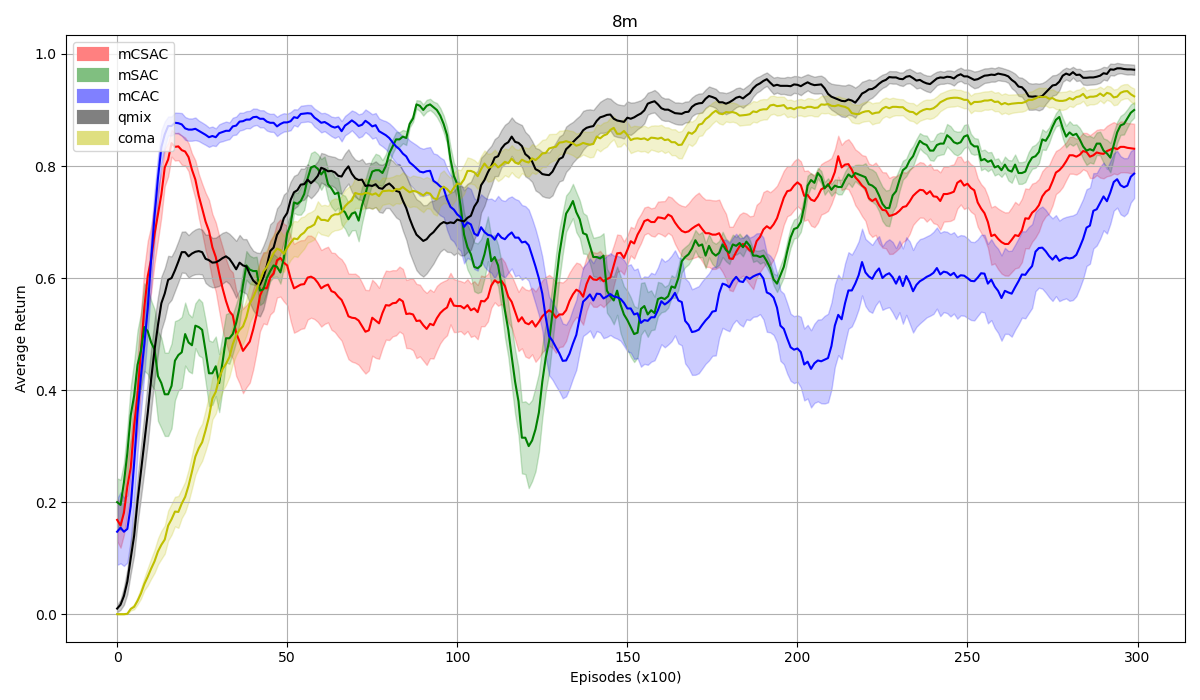} }}%
	\qquad
	\subfloat[\centering $3m$]{{\includegraphics[width=5cm]{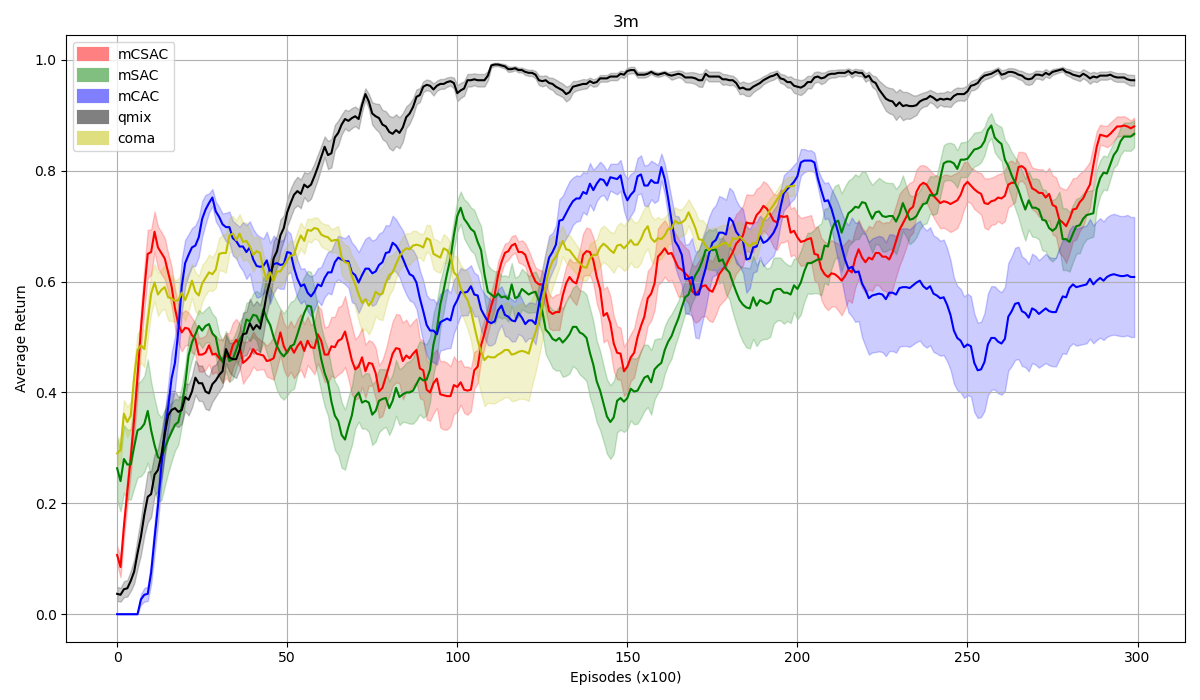} }}%
	\caption{The performance curves for mCSAC, mSAC, mCAC, and QMIX, COMA on different StarCraft II micromanagement combat maps with homogeneous agents.}%
	\label{fig:example}%
\end{figure}
\begin{figure}[!t]%[htbp]
	\centering
	\subfloat[\centering $1c3s5z$]{{\includegraphics[width=5cm]{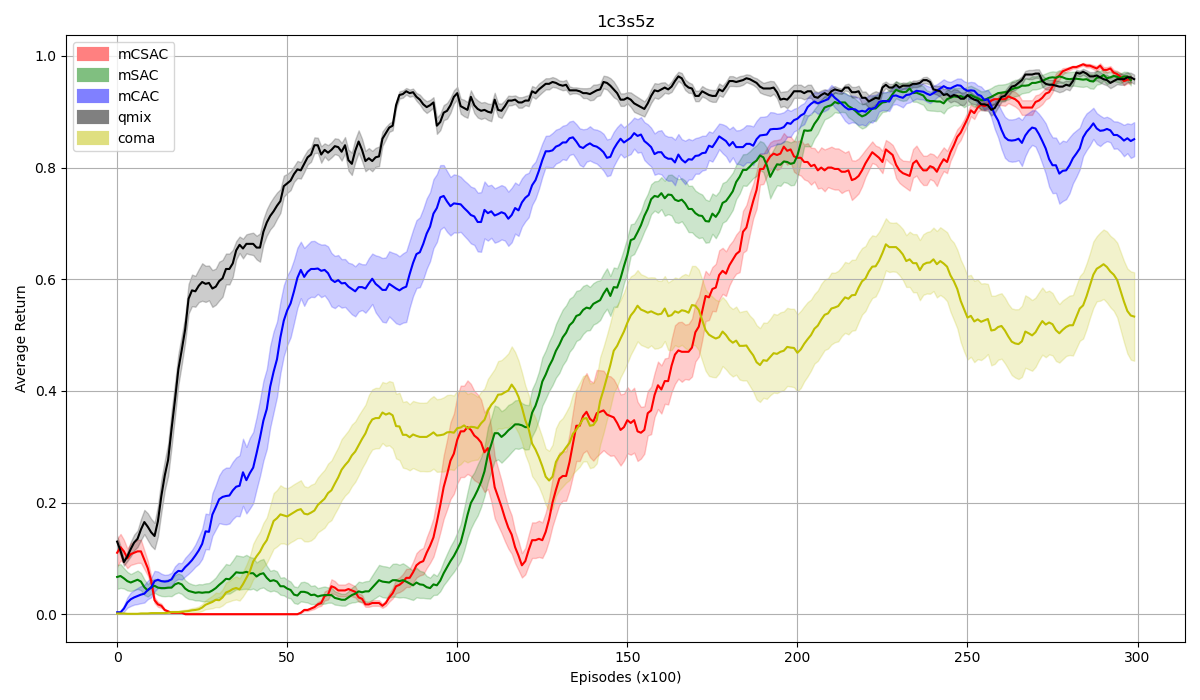} }}%
	\qquad
	\subfloat[\centering $3s5z$]{{\includegraphics[width=5cm]{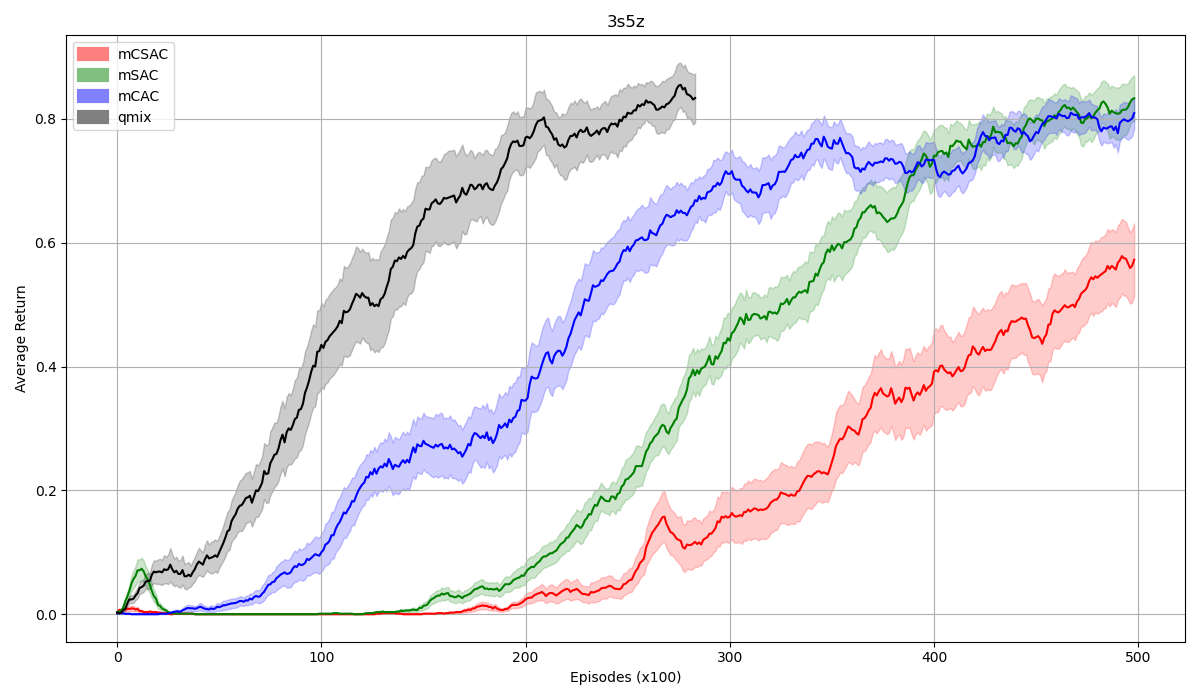} }}%
	\qquad
	\subfloat[\centering $3s\_vs\_5z$]{{\includegraphics[width=5cm]{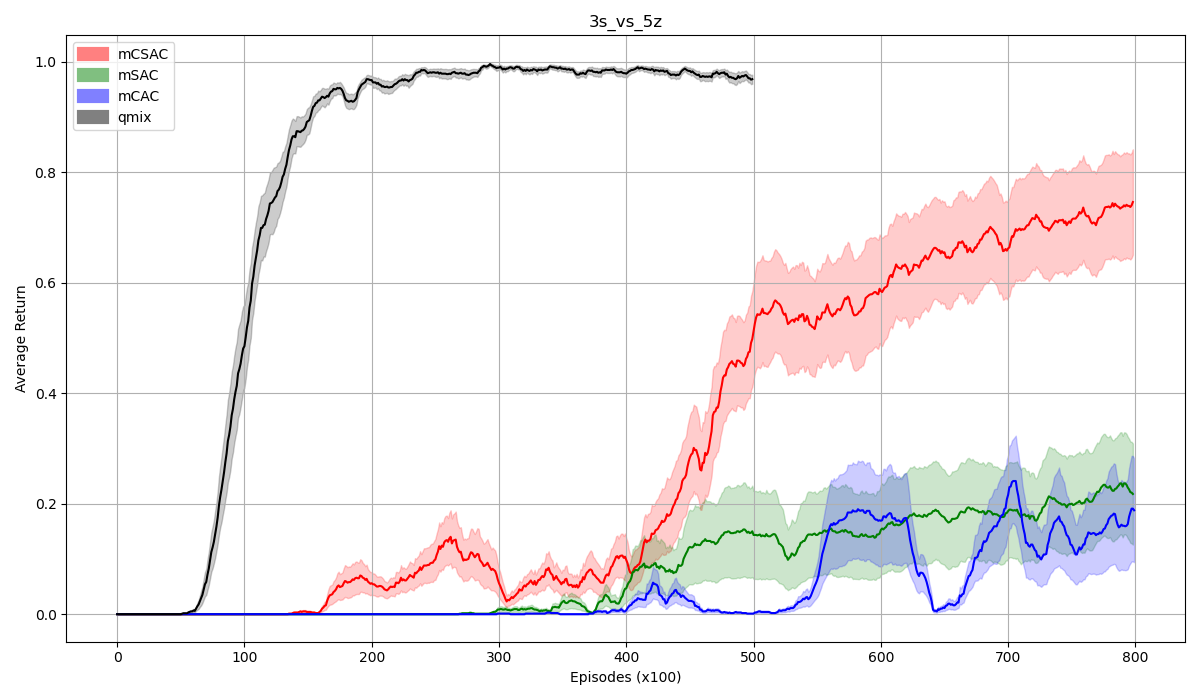} }}%
	\caption{The performance curves for mCSAC, mSAC, mCAC, and QMIX, COMA on different StarCraft II micromanagement combat maps with heterogeneous agents. In $3s5z$ and $3s\_vs\_5z$, COMA method achieves zero test win-rate according to the experimental results in \cite{b18}, so we don't plot the COMA curves in these graphs. }%
	\label{fig:example}%
\end{figure}
\begin{figure}[!t]%[htbp]
	\centering
	\subfloat[\centering $2c\_vs\_64zg$]{{\includegraphics[width=5cm]{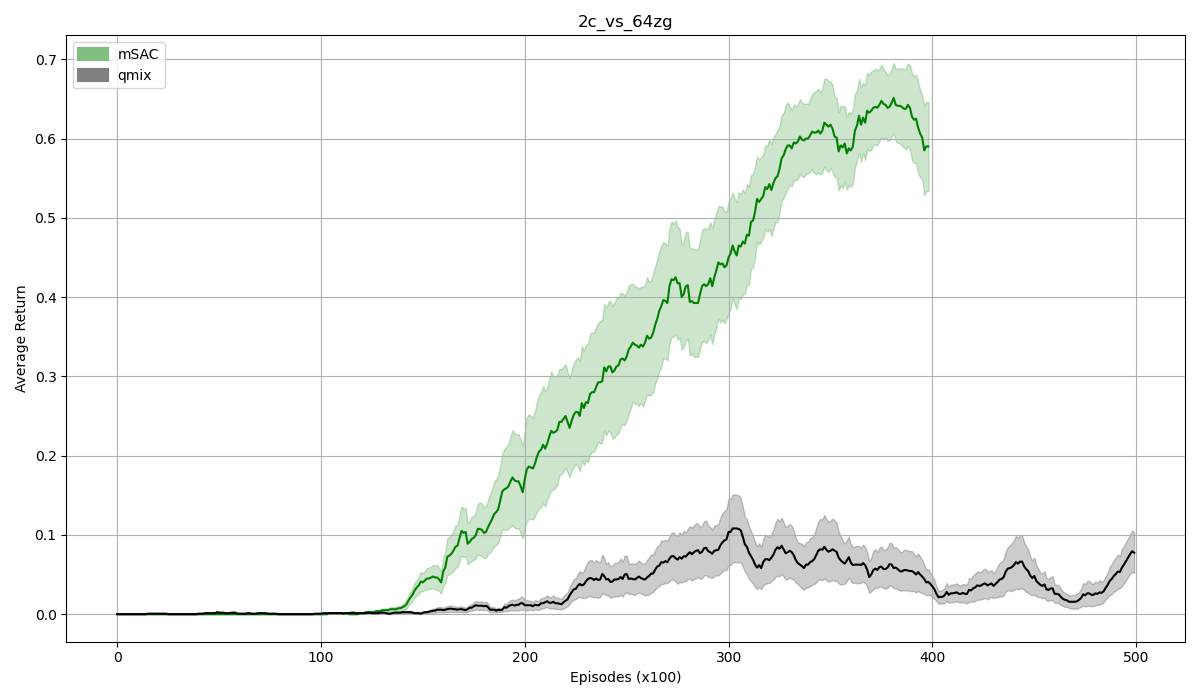} }}%
	\qquad
	\subfloat[\centering $MMM2$]{{\includegraphics[width=5cm]{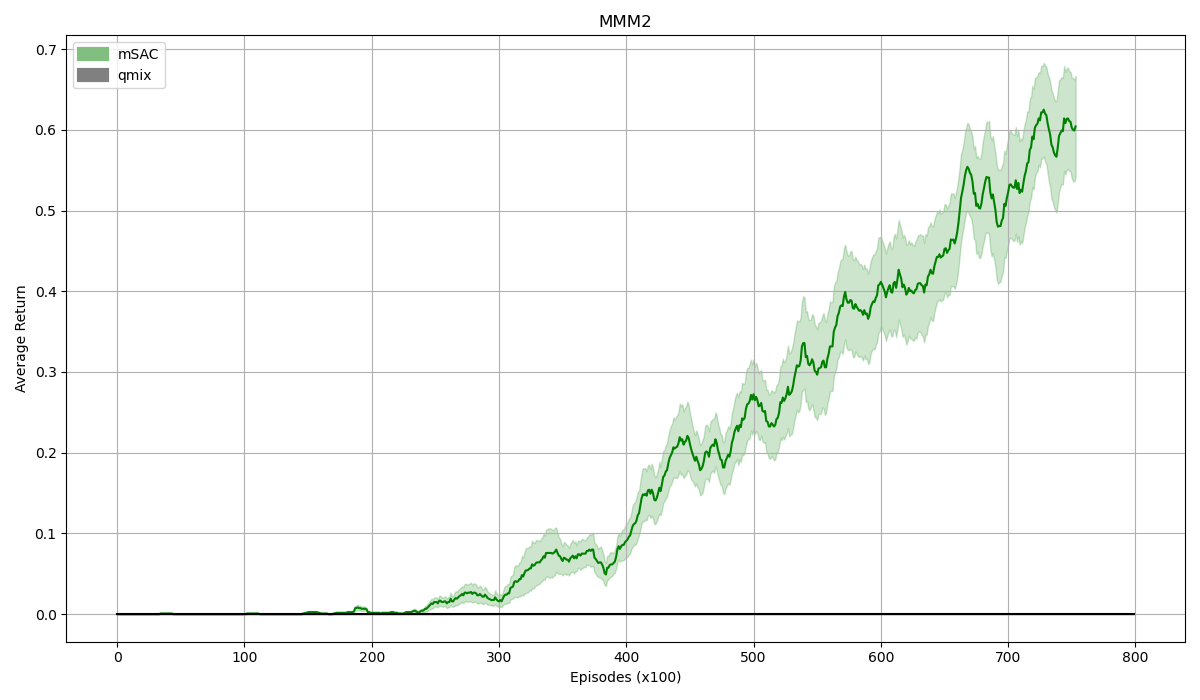} }}%
	\qquad
	\subfloat[\centering $bane\_vs\_bane$]{{\includegraphics[width=5cm]{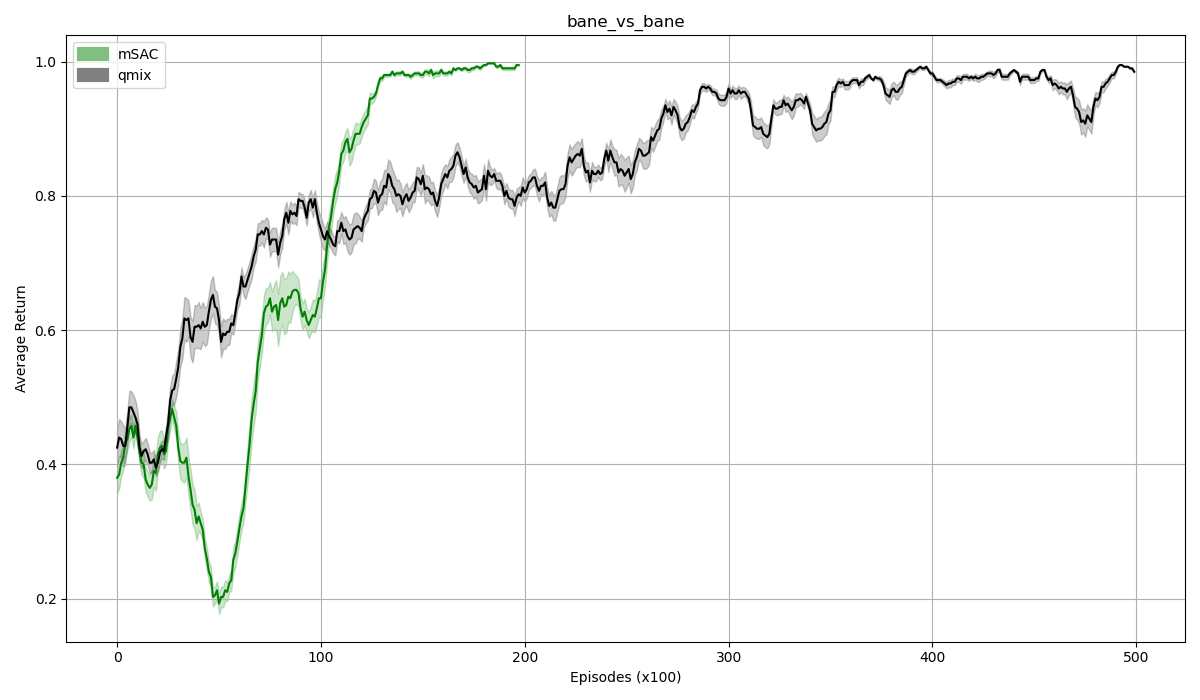} }}%
	\caption{The performance curves for mSAC and QMIX on large action space tasks: $2c\_vs\_64zg$, $MMM2$ and $bane\_vs\_bane$. }%
	\label{fig:example}%
\end{figure}
% Our experiments reveal that CAC outperforms, or is competitive with, all of its ablations. analyze
In this part, we compared the performance of our method mSAC, its variant methods mCSAC, mCAC, and advanced Qmix and COMA methods in the different maps, including
% aforementioned discrete tasks. 
 maps with homogeneous agents ($3m$, $8m$), and maps with heterogeneous agents ($1c3s5z$, $3s5z$, $3s\_vs\_5z$), maps that agent contains a large action space ($2c\_vs\_64zg$, $MMM2$, $bane\_vs\_bane$, $27m\_vs\_30m$).
 The performance test win rate learning curves are shown in Figure 3, 4 and 5 respectively.
 % in the above discrete task
%Tested the performance of the mSAC method,  
Through the experimental results, it can be found that the method mSAC proposed in this chapter is similar to  policy-based method COMA on the map of homogeneous agents, and is significantly better than COMA on other maps. In the maps $8m$, $1c3s5z$, $3m$, $3s5z$, compared with the current value-based method Qmix, it has similar asymptotic performance. In the map with a large action space for the agent ($2c\_vs\ _64zg$, $MMM2$, $bane\_vs\_bane$), its performance is significantly better than Qmix. In some relatively difficult tasks, such as $3s\_vs\_5z$, the performance of all policy-based methods is worse than Qmix but not much.
After carefully analyzing the experimental results, the following observations and conclusions can be drawn:
% We found that our proposed multi-agent decomposed soft actor-critic method mSAC obtained competitive asymptotic performance with current state-of-the-art value-based method Qmix in 8m, 1c3s5z, 3m
% % (these results are deferred in the Appendix), 
% 3s5z,  and outperforms Qmix marginally in large-action space map, such as $2c\_vs\_64zg$, $MMM2$ and $bane\_vs\_bane$.
% And in some relatively hard tasks, like map $3s\_vs\_5z$, the performance of all policy gradient methods are worse than Qmix but not much. 

% (Note that in order to satisfy equation (1) all variant methods use only one linear layer mixing network but the Qmix method utilizes two layer non-linear mixing network, so there are certain prior advantages for Qmix while comparing results with our proposed methods.)

% From these results, we can draw the following observations and conclusions:
{\bf 1. Soft policy iteration paradigm is also effective in multi-agent scenarios. }
From the comparison of the results of all maps we found that mCAC behaves worse than the other methods both in stability and asymptotic performance, which indicates that soft policy iteration paradigm is usually beneficial to the robust policy improvement in multi-agent policy gradient setting. %As in \cite{b11}, 
We conjecture that this is because simultaneously maximizing expected return and entropy can make the agent explore more widely and efficiently, and can capture multiple modes of near-optimal behaviours. 
% the overall performance decreases when we turn off soft q values

{\bf 2. It is important to jointly optimize the entire policy distribution On tasks where the agent has a relatively large action space. }
% In large action space map like $2c\_vs\_64zg$ 
% (the same analysis can be conducted for map $MMM2$ and $bane\_vs\_bane$)
% For example, in map $2c\_vs\_64zg$ 
% , the Colossi units have large action space $|A|$=70. In Qmix method, the agents execute in a decentralized way, and they only choose action greedily according to its local Q values. The additional exploration mechanism is achieved only by the $\epsilon$-greedy paradigm, which is not very efficient in exploring individual agents’ behavior space.
% While in our proposed method mSAC, each agent executes action according to its individual policy that is a learned categorical distribution, and optimize the whole probabilistic distribution jointly to maximize the sum of expected returns and policy entropy.
% Intuitively, this is more reasonable than Qmix. This analysis may give us some insights on how to solve large action space tasks in the future
For example, in the map $2c\_vs\_64zg$, the Colossi unit has a large action space $|A|=70$. In the Qmix method, all agents are executed in a decentralized manner. Each agent greedily selects actions based on its local action value function. In a certain state, there is only one action that maximizes the local action value, and the others actions are given the same selection probability $\epsilon$. At a certain time, in map $2c\_vs\_64zg$, the probability of ally unit attacking a specific enemy among all the 64 enemy units is very high, while the The probability of attacking the other 63 units is the same $\epsilon$.
%which is often unreasonable, because at a certain time there may be many actions that are optimal at the same time, and the $\epsilon$-greedy paradigm cannot realize the guided exploration function.
% As a result, Qmix's exploration mechanism can only be realized, which makes it difficult to effectively explore different areas of the agent's action space.
While in the mSAC method, each agent executes actions according to its own strategy, that is, the learned categorical distribution, and can choose different areas of the action space in a planned way.
By jointly optimizing the entire probability distribution to maximize the sum of expected returns and strategy entropy, intuitively speaking, this is more reasonable and effective than Qmix's $\epsilon$-greedy paradigm exploration on tasks with large action spaces.

{\bf 3. Counterfactual advantage functions are not always effective, and are more important in relatively complex tasks. }
For easy environments, like map $8m, 1c3s5z, 3s5z$, the performance of mCSAC and mSAC is similar, but in harder environment, like map $3s\_vs\_5z$ , the performance gap of mCSAC and mSAC is larger, which indicates that attribution of global reward is critical for solving this harder task, the counterfactual advantage function partially addresses the issue, could gradually learn a reasonable credit assignment during training in some tasks but is not always effective. Moreover, we carefully analyzed the performance of each seed and found that after training of 80000 episodes, in some seeds the test win rate can perform up to 90\%, and some seeds are zero. We speculate that this may because it's more difficult to explore good strategies in difficult maps, which indicates that effective exploration would be a important research problem.

% \subsection{ Continuous action spaces: Multiple Partial Environment}
% \subsubsection{[]} 
% \subsubsection{Results}

\section{Conclusion and Future Works}

In this paper, we presented the new decomposed multi-agent soft actor-critic method (mSAC) that incorporates value function decomposition, soft policy iteration, and counterfactual advantage function (optional), which supports efficient off-policy learning and addresses the issue of credit assignment partially.
% in both discrete and continuous action spaces potentially. 
mSAC learns the distributional policy for each agent simultaneously which seems like a guided distributional exploration implicitly,
% guide individuals’ behaviors,
which is especially important in large action space task through the experimental results. 

In addition, we empirically investigate the performance of mSAC and its variant methods
% and study the effects of the different main components 
in StarCraft II micromanagement cooperative multi-agent benchmark. Experimental results demonstrate that mSAC can achieve relatively stable and efficient multi-agent off-policy learning and outperforms, or is competitive with, current main policy-based algorithms and value-based approaches (e.g. COMA, and Qmix) on most tasks, and achieves very good results in large action space task like \text {\em 2c\_vs\_64zg} and $MMM2$.

However, in this paper, we only study the effect of counterfactual multi-agent soft actor-critic paradigm on the discrete domain. 
Experiments under continuous domain need to be studied. In addition,
The more solid theoretical analysis of the algorithm will be needed, and at the same time, how to explore more efficiently will be a valuable future work. 
%And another widely open avenue for future work lies in improving the sample efficiency, to allow %applications to domains that do not easily allow fast simulation at similar scales.

%\section*{Acknowledgment}
%
%Thanks for insightful discussion with Rui Yang, Weiming Liu, Xin Yao and Shaochen Wang. The
%research is partially supported by the National Natural
%Science Foundation of China under grant No.U19B2044
%and No.61836011.
%%\section{References}

\section*{APPENDIX}

% h llllll
\begin{table*}[!htbp]
	\centering
	\caption{StarCraftII Micromanagement Maps Parameters}
	 \scalebox{0.8}{
	\begin{tabular}{llllll}
		\hline
		Name & Ally Units & Enemy Units  &  Episode Length & Obs.  Dim & Action Dim \\
		\hline
		Easy\\
		\hline
		3m & 3 Marines& 3 Marines  & 60& 30&9 \\
		8m  & 8 Marines& 8 Marines &120 &80& 14 \\
		1c3s5z  & 1 Colossi, 3 Stalkers, 5 Zealots& 1 Colossi, 3 Stalkers,5 Zealots&180 &162& 15\\
		bane\_vs\_bane & 4 Banelings,20 Zerglings&4 Banelings,20 Zerglings&200&336& 30\\
		\hline
		Hard\\
		\hline
		3s5z & 3 Stalkers, 5 Zealots& 3 Stalkers, 5 Zealots&150&128& 14\\
		3s\_vs\_5z  & 3 Stalkers& 5 Zealots & 250&48& 11\\
		2c\_vs\_64zg  & 2 Colossi &64 Zerglings &400& 332 &70 \\
		10m\_vs\_11m  & 10 Marines&11 Marines& 150& 105&17\\
		\hline
		SuperHard\\	
		\hline
		27m\_vs\_30m & 27 Marines&30 Marines& 180&285& 36\\
		3s5z\_vs\_3s6z&3 Stalkers,5 Zealots&3 Stalkers,6 Zealots&170&136&15\\
		MMM2  &1 Medivac,2 Marauders,7 Marines&1 Medivac,3 Marauders,8 Marines&180 &176 &18\\
		\hline
	\end{tabular}}
	%	\caption{Latex default table}
	%\label{tab:plain}
	\label{tab:booktabs}
\end{table*}

{\bf Some Equation Proof Details}

The important equation for efficiently calculating the expectation of the joint Q values using the expectation of the local Q values following local policies:
\begin{equation}
	\begin{aligned}
	&\mathbb{E}_{\boldsymbol{\pi}}\left[Q^{t o t}(\boldsymbol{s},\boldsymbol{\tau}, \boldsymbol{a})\right]
	&=\sum_{i} k^{i}(\boldsymbol{s}) \mathbb{E}_{{\pi}^{i}} \left[ q^{i}\left(\boldsymbol{\tau}^{i}, a^{i}\right)\right]+ b(\boldsymbol{s})\\
	&=q^{mix}(\boldsymbol{s},\mathbb{E}_{{\pi}^{i}} \left[ q^{i}\left(\boldsymbol{\tau}^{i}, a^{i}\right)\right])
	\end{aligned}
\end{equation}

The detailed proof is as follows:
\begin{equation}
	\begin{aligned}
		&\mathbb{E}_{\pi}
		\left[
		Q^{t o t}(
		\boldsymbol{s}, \boldsymbol{\tau}, \boldsymbol{a}
		)
		\right]\\
		&=\sum_{\boldsymbol{a}}
		\pi( \boldsymbol{a} | \boldsymbol{\tau} )
		Q^{t o t}(
		\boldsymbol{s}, \boldsymbol{\tau}, \boldsymbol{a}
		)
		\\
		&=\sum_{\boldsymbol{a}}
		\pi( \boldsymbol{a} | \boldsymbol{\tau} )
		\left[
		\sum_{i}
		k^{i}(\boldsymbol{s})
		q^{i}
		\left(
		\boldsymbol{\tau}^i, a^i
		\right)
		+b(\boldsymbol{s})
		\right] 
		\\
		&=\sum_{\boldsymbol{a}}
		\pi(\boldsymbol{a} | \boldsymbol{\tau})
		\sum_{i}
		k^{i}(\boldsymbol{s})
		q^{i}
		\left(
		\boldsymbol{\tau}^{i}, a^{i}
		\right)
		+\sum_{\boldsymbol{a}}
		\pi(\boldsymbol{a} | \boldsymbol{\tau})
		b(\boldsymbol{s})
		\\
		&=\sum_{i}
		\sum_{\boldsymbol{a}}
		\pi(\boldsymbol{a} | \boldsymbol{\tau})  
		k^{i}(\boldsymbol{s}) 
		q^{i}
		\left(
		\boldsymbol{\tau}^{i}, a^{i}
		\right)
		+ b(\boldsymbol{s})
		\\
		& \text{below omit } b(\boldsymbol{s}) \text{ for simplicity} \\
		&=\sum_{i} k^{i}(\boldsymbol{s}) \sum_{\boldsymbol{a}} \boldsymbol{\pi}(\boldsymbol{a} | \boldsymbol{\tau})   q^{i}\left(\boldsymbol{\tau}^{i}, a^{i}\right)\\
		&=\sum_{i} k^{i}(\boldsymbol{s}) \sum_{\boldsymbol{a}} {\pi}^{i}({a^i} | \boldsymbol{\tau}^{i}) \boldsymbol{\pi}^{-i}(\boldsymbol{a}^{-i} | \boldsymbol{\tau}^{-i}) q^{i}\left(\boldsymbol{\tau}^{i}, a^{i}\right)\\
		&=\sum_{i} k^{i}(\boldsymbol{s}) \sum_{\boldsymbol{a^i}}  {\pi}^{i}({a^i} | \boldsymbol{\tau}^{i}) q^{i}\left(\boldsymbol{\tau}^{i}, a^{i}\right) \sum_{\boldsymbol{a^{-i}}} \boldsymbol{\pi}^{-i}(\boldsymbol{a}^{-i} | \boldsymbol{\tau}^{-i})\\
		&=\sum_{i} k^{i}(\boldsymbol{s}) \sum_{\boldsymbol{a^i}} {\pi}^{i}({a^i} | \boldsymbol{\tau}^{i}) q^{i}\left(\boldsymbol{\tau}^{i}, a^{i}\right)\\
		&=\sum_{i} k^{i}(\boldsymbol{s}) \mathbb{E}_{{\pi}^{i}} \left[ q^{i}\left(\boldsymbol{\tau}^{i}, a^{i}\right)\right]
	\end{aligned}
\end{equation}

\begin{equation}
	\begin{aligned}
		&\mathbb{E}_{\boldsymbol{\pi}}\left[Q^{t o t}(\boldsymbol{s},\boldsymbol{\tau}, \boldsymbol{a}) -\alpha \log \boldsymbol{\pi} \left(\boldsymbol{a} \mid \boldsymbol{\tau} \right) \right]\\
		&=\sum_{i} k^{i}(\boldsymbol{s}) \mathbb{E}_{{\pi}^{i}} \left[ q^{i}\left(\boldsymbol{\tau}^{i}, a^{i}\right)\right]+ b(\boldsymbol{s})+ \alpha  H(\boldsymbol{\pi}) \\
		&=q^{mix}(\boldsymbol{s},\mathbb{E}_{{\pi}^{i}} \left[ q^{i}\left(\boldsymbol{\tau}^{i}, a^{i}\right)\right])+ \alpha H(\boldsymbol{\pi})
	\end{aligned}
\end{equation}
we could approximate the above equation using following equation,

\begin{equation}
	\begin{aligned}
		&q^{mix}(\boldsymbol{s},\mathbb{E}_{{\pi}^{i}} \left[ q^{i}\left(\boldsymbol{\tau}^{i}, a^{i}\right)  -\alpha \log {\pi}^{i} \left({a^i} \mid \boldsymbol{\tau}^i \right) \right])\\
		&=\sum_{i} k^{i}(\boldsymbol{s}) \mathbb{E}_{{\pi}^{i}} \left[ q^{i}\left(\boldsymbol{\tau}^{i}, a^{i}\right)\right]+
		\sum_{i} k^{i}(\boldsymbol{s}) \mathbb{E}_{{\pi}^{i}} \left[- \alpha \log \pi^{i}\left(a^{i} | \boldsymbol{\tau}^{i}\right)\right]
		+b(\boldsymbol{s})\\
		&=\sum_{i} k^{i}(\boldsymbol{s}) \mathbb{E}_{{\pi}^{i}} \left[ q^{i}\left(\boldsymbol{\tau}^{i}, a^{i}\right)\right]+ b(\boldsymbol{s})+ \alpha \sum_{i} k^{i}(\boldsymbol{s}) H(\boldsymbol{\pi}^i) 
	\end{aligned}
\end{equation}

if we use a different mixing network for entropy term, the equation becomes:
\begin{equation}
	\begin{aligned}
		&q^{mix1}(\boldsymbol{s},\mathbb{E}_{{\pi}^{i}} \left[ q^{i}\left(\boldsymbol{\tau}^{i}, a^{i}\right)  \right]) + q^{mix2}(\boldsymbol{s},\mathbb{E}_{{\pi}^{i}} \left[-\alpha \log {\pi}^{i} \left({a^i} \mid \boldsymbol{\tau}^i \right) \right])\\
		&=\sum_{i} k^{i}_{1}(\boldsymbol{s}) \mathbb{E}_{{\pi}^{i}} \left[ q^{i}\left(\boldsymbol{\tau}^{i}, a^{i}\right)\right]+ b(\boldsymbol{s})+ \alpha \sum_{i} k^{i}_{2}(\boldsymbol{s}) H(\boldsymbol{\pi}^i) 
	\end{aligned}
\end{equation}

\vspace{12pt}
\end{document}